# Digital measurement of droplet flame diameter in microgravity combustion images using Segment Anything Model 2 with automatic prompt selection

Minghui Xu[a], Chaoyi Zhou[b], Aaron P. Cecil[a], Xi Liu[a], Siyu Huang[b], Yuhao Xu[a,*]

[a]*Department of Mechanical Engineering, Clemson University, Clemson, SC 29634, USA*
[b]*Visual Computing Division, School of Computing, Clemson University, Clemson, SC 29634, USA*

---

**Abstract**

Flame diameter is a key measurable parameter in microgravity droplet combustion, but its extraction from self-illuminated frames remains difficult because soot tails, blurred luminous boundaries, chamber reflections, and droplet drift introduce substantial measurement bias and operator dependence. This work presents an AI-enabled digital measurement workflow for automated flame diameter from combustion images. The workflow integrates automatic prompt-point generation into Segment Anything Model 2, employing Random Sample Consensus (RANSAC)-based circle fitting. The automatic prompt strategy removes subjective manual point selection, while the video memory mechanism maintains temporal consistency for drifting droplets, and the RANSAC fitting rejects soot-tail pixels as geometric outliers.

The method is validated by 19,537 flame images of *n*-heptane, *n*-decane, and *n*-octane droplets with varying initial diameters. Compared with manual-reference measurements, the proposed workflow achieves a mean relative agreement of 96.9%, a mean absolute percentage error of 3.1%, and substantially outperforms conventional Hough circle detection, which performed worse under the same evaluation conditions. The results also show that the measurement accuracy improves with increasing droplet size. The proposed workflow has a combined standard uncertainty of 8.54% and achieves approximately a 229-fold improvement in efficiency over manual measurement.

These results demonstrate that the proposed SAM2-based workflow provides a reproducible, fully automated, and metrologically characterized digital measurement system for extracting flame diameter from challenging combustion images. The approach supports high-throughput combustion diagnostics and illustrates that AI-based segmentation can be integrated into quantitative measurement workflows for digitalized image-based metrology.

---

## 1. Introduction

The droplet combustion of liquid fuels serves as a canonical configuration for studying the combustion of various liquid fuels, capturing multiphase effects, moving boundaries, unsteady liquid and gas transport, soot formation, and flame dynamics. This configuration is often achieved by conducting experiments in a microgravity environment [1-11] aboard the International Space Station (ISS). Under microgravity conditions, the absence of buoyancy-induced convection results in a spherically symmetric flame structure, providing an ideal environment for investigating intrinsic combustion and soot formation mechanisms and generating valuable experimental data for developing and validating detailed numerical simulation models [1, 3, 9].

Accurate measurements of flame diameter ($D_f$) in fuel droplet combustion under microgravity are critical for understanding flame dynamics, developing theoretical models for predicting flame temperature [4, 5], and providing experimental data for validating numerical models [1, 3, 9]. Despite its significance, precisely determining flame diameters in an automatic fashion remains a challenge, particularly under sooty conditions where flame boundaries become blurry and distorted. These difficulties compromise the reliability of automated diameter extraction and can introduce systematic bias in experimental measurements. This work aims to overcome these difficulties and develop an automated method for measuring flame diameter from video frames, while ensuring the accuracy of these measurements.

### *1.1 Problem description*

Figure 1 shows representative self-illuminated flame images of burning *n*-heptane droplets with two different initial droplet diameters ($D_0$ =1.93 mm, $D_0$ = 3.47 mm) at the early, mid, and end stages of combustion. The selected images in Figure 1 are to highlight several non-ideal image conditions that complicate the automated measurement process for flame diameters relative to an ideal, sharply bounded circle. In the early stage (Figure 1a and 1d), a soot-induced annular region can obscure the actual flame boundary. Accurate detection of this obscure annulus is necessary for precise measurement of $D_f$. In this situation, the correct detection of the flame boundary becomes challenging to detect because of "corona effects" (halo-like optical artifacts caused by diffraction and camera blooming when the luminous soot core is saturated). These corona effects produce stray-light halos with contrast similar to the true annular boundary, which can cause false edge detection. This issue must be addressed to obtain accurate flame measurements [5].

Figure 1b shows a bright, yellow flame with a soot tail at the mid-combustion stage. The soot tail will undoubtedly negatively affect the automatic measurements of $D_f$. This imposes another challenge to overcome, as the soot tail should not be considered as part of the flame circle when measuring the flame diameter. In contrast, Figure 1e presents a relatively clean bright flame without a soot tail, which is relatively easy to measure because of the clear flame boundary against the image background.

During microgravity droplet combustion experiments, droplets can drift around if they are untethered as an artifact of the complex experiments [1, 3-5]. As a result, the droplet flame would drift, as shown in Figure 1c, which leads to the shifting of the centroid and then causes the loss of target identification.

Figure 1f shows an example of a spherical flame with a blue boundary. The blue color is a result of excited CH radicals present in the high-temperature zone [12]. The diameter of the blue flame is relatively straightforward to measure, as the boundary is clearly defined against the image background. Note that blue flames generally only occur in large droplets (around $D_0 > 2$ mm).

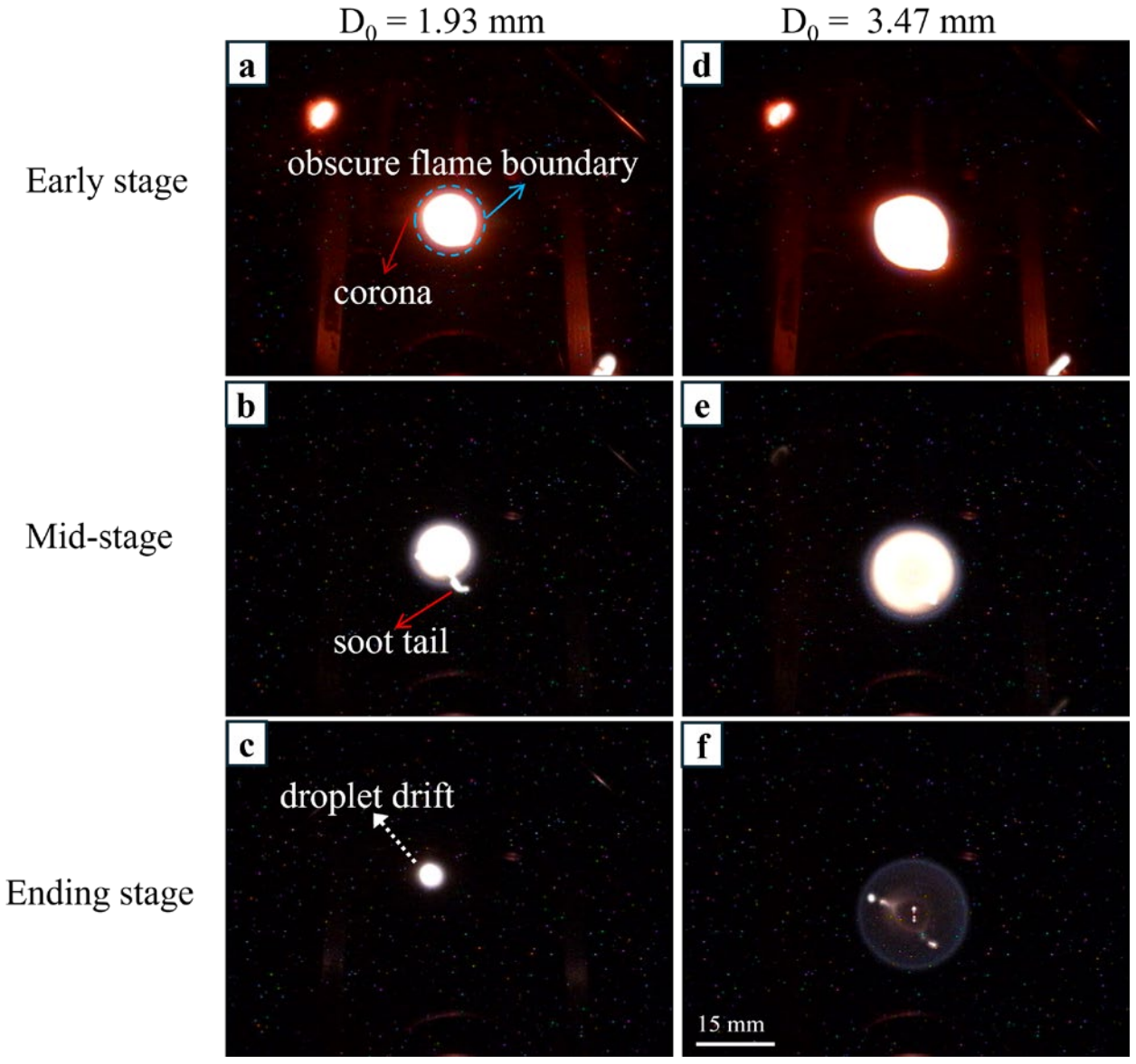

**Figure 1.** Representative self-illuminated flame images [13] of burning *n*-heptane droplets with two different initial droplet diameters ($D_0$ = 1.93 mm, $D_0$ = 3.47 mm) at various combustion stages. **a. & d.** Obscured flame boundary with luminous corona. **b.** A soot tail extending from the main flame zone. **c.** Droplet drift during experiments. **e.** A bright, yellow flame without a soot tail. **f.** A blue flame for a larger droplet.

### *1.2 Related previous work*

Circle measurements play a crucial role in various scientific and engineering applications, including the inspection of mechanical components, optical diagnostics, medical imaging, and droplet combustion. This section summarizes relevant prior work in circle detection. The Hough circle transform is the most conventional method for circle detection [14, 15], e.g., More and Patil proposed a robust image-based method to detect and measure the circles of industrial components, like a water pump pulley, using the conventional circular Hough transform [16]. Lee et al. integrated the Segment Anything Model (SAM) into a pipeline for vision-based geometric measurement of circular cross-sections (e.g., stem diameters), combining segmentation, mask refinement, skeletonization, and diameter estimation [17]. Heo et al. studied aortic diameter measurement, using a deep learning (DL) segmentation method combined with ellipse calibration to align automatic diameter outputs with the clinical gold standard [18]. Rasul et al. developed a dynamic differential image circle diameter measurement method for droplet diameter measurements, applying gradient-based boundary detection and least-squares circle fitting to time-series droplet images [19].

Traditional methods for flame diameter measurements often rely on manual image processing techniques or threshold-based segmentation methods [1, 4, 5]. Although these methods can provide approximate results, they have some limitations. First, the presence of soot tails introduces irregular bright regions that bias the diameter fit and compromise repeatability. Moreover, manual measurements are subjective and require considerable manpower, as typically several hundred images are available for each set of combustion experiments [3, 4, 8, 10, 20-23]. Several studies measured the $D_f$ by manually positioning a virtual ellipse around the outer blue zone of the bright central core, using Image-Pro software to enhance the obscure blue boundary [3-5, 8, 10, 19, 22]. There was no prior work reporting the automated measurement of flame images recorded during droplet combustion experiments in which droplets are untethered and can therefore drift.

### *1.3 Novelty and overview of this work*

The key novelties of this work are discussed below. To overcome these limitations, a fully automated digital measurement workflow is developed to measure flame diameter from images recorded during combustion, built upon Segment Anything Model 2 (SAM2) segmentation and enhanced by domain-aware strategies. First, an automatic symmetric prompt-generation algorithm is incorporated into SAM2 to eliminate the subjectivity of manual prompt selection and improve measurement reproducibility. An accurate fitting method, RANSAC, is also employed to discard the soot tail points as outliers and then refit the inliers using geometric least squares. Furthermore, the droplet-drifting issue is resolved by SAM2's video model, ensuring frame-by-frame consistency and allowing smooth diameter curves even when the droplet flame is drifting. The average accuracy of flame diameter measurements using this new framework is 96.9%, showing a significant improvement over the traditional circular Hough transform (55.9%). The performance of the workflow is validated across 19,537 images for three fuels, including *n*-heptane, *n*-decane and *n*-octane, with varying initial diameters and uncertainty-related analysis. Measurement accuracy increases with increasing initial droplet diameter. The proposed workflow achieved approximately a 229-fold improvement in measurement efficiency, requiring only a small fraction of the processing time required for manual measurements. This newly developed method successfully integrates advances from SAM2-based diameter measurement into the specialized task of combustion imaging, achieving accurate, repeatable, and automated flame-diameter extraction under challenging conditions.

## 2. Experimental method

The droplet combustion experiments were conducted remotely via access to the Multi-user Droplet Combustion Apparatus (MDCA) onboard the ISS. The experiments were part of the 'surrogate fuel' component of the Flame Extinguishment Experiment-2 (FLEX-2) program on the ISS [3-5, 24]. The experimental method and apparatus are briefly discussed below. Detailed information about the experimental hardware can be found in [25, 26].

The fuel droplet was formed by dispensing liquid through two closely spaced needles, creating a liquid bridge. When the desired droplet size was achieved, the droplet was detached by retracting the needles, leaving an untethered droplet behind. The droplet was then ignited by energizing electrical coils positioned on opposite sides of the droplet, which were immediately retracted after ignition, minimizing disturbance to the flame. All tests were performed in room-temperature air at atmospheric pressure (1 atm). The self-illuminated flame images used in this work were captured using a color camera with a resolution of 0.3 MP (640×480 pixels) at 30 frames per second (fps). Backlit images of the droplets were captured simultaneously with the flame images using a High Bit-depth Multispectral (HiBMs) camera operating at 30 fps. The backlit images are not used in this paper, as the focus of this work is not on analyzing droplet diameters; however, they are mentioned for completeness regarding the experimental hardware.

Flame images of *n*-heptane are used to develop the framework in this study. Subsequently, the flame diameters of *n*-heptane, *n*-decane, and *n*-octane are measured using this new framework and compared with manual-reference measurements (m-measurements), which were reported previously [5], to verify its capabilities and compatibility. Flame images of *n*-heptane, *n*-decane, and *n*-octane are downloaded from a website maintained by NASA that promotes open access to research data [13].

## 3. Digital flame diameter measurement framework

Recently, the Segment Anything Model 2 has demonstrated remarkable generalization capability in object segmentation across diverse image domains [27, 28]. Unlike conventional methods, SAM2 uses an interactive, prompt-based mechanism that enables accurate boundary detection even under challenging conditions. The core architecture of SAM2 consists of three principal modules: an image encoder, a prompt encoder (e.g., points, bounding boxes, or masks), and a mask decoder. The image encoder extracts feature embeddings from the original input image, while the prompt encoder transforms the prompt information into prompt embeddings. Both embeddings are fused with the mask decoder to generate high-quality binary segmentation masks. A significant improvement in SAM2 lies in the hierarchical vision backbone (Hiera), which provides a stronger feature representation than the vision transformer used in SAM. Furthermore, SAM2 includes a memory encoder and a memory attention module. The memory encoder combines the features of the current frame with the past segmentation results. The target object is then segmented by the memory attention module based on temporal information from previous frames. This design enables SAM2 to consistently segment target objects across different sequences, thereby reducing frame-to-frame variation and minimizing temporal drift.

Despite SAM2 achieving remarkable success in vision benchmarks, it has not been used to analyze combustion images and obtain quantitative measurements. In this work, we extend SAM2 to extract flame boundaries from droplet combustion images with blurred edges and the influence of soot particles, thereby enabling accurate flame diameter measurements that can be applied to different liquid fuels. We use the publicly released Segment Anything Model 2 implementation from the official repository [27]. Specifically, we employ the SAM 2.1 checkpoint (sam2.1_hiera_tiny.pt) with its corresponding configuration file, and keep the default input setting unless otherwise noted. We do not fine-tune SAM2 on our data. The model weights are pretrained by the original authors; thus, the training data and procedure are identical to those of Ravi et al. [27]. Figure 2 presents the complete flowchart for the automatic measurement of flame diameter. To get a correct segmentation for the flame boundary, the four boundary prompt points (p-points, for estimating the flame boundary) and one centroid p-point of the droplet (for locating the center of the flame) are first extracted using a ray-contour intersection method, only for the first frame of a sequence of combustion flame images. Here, the 'contour' refers to a coarse flame boundary obtained from the lightweight HSV/brightness mask described in Section 3.3 and is used only to initialize p-points for the first frame; SAM2 then generates the final segmentation mask. These p-points are automatically selected as symmetric p-points and fed into SAM2, where the image features are extracted, the p-points are encoded, and a segmentation mask is generated. Meanwhile, a memory attention module is employed to ensure the temporal consistency across the whole image sequence. Eventually, the segmented flame boundary is fitted by a circle solver (see the details in Section 3.1), and the quantitative flame diameter is obtained in pixels, which is later converted to millimeters after applying the calibration factor (640 pixels = 93.2 mm [29]).

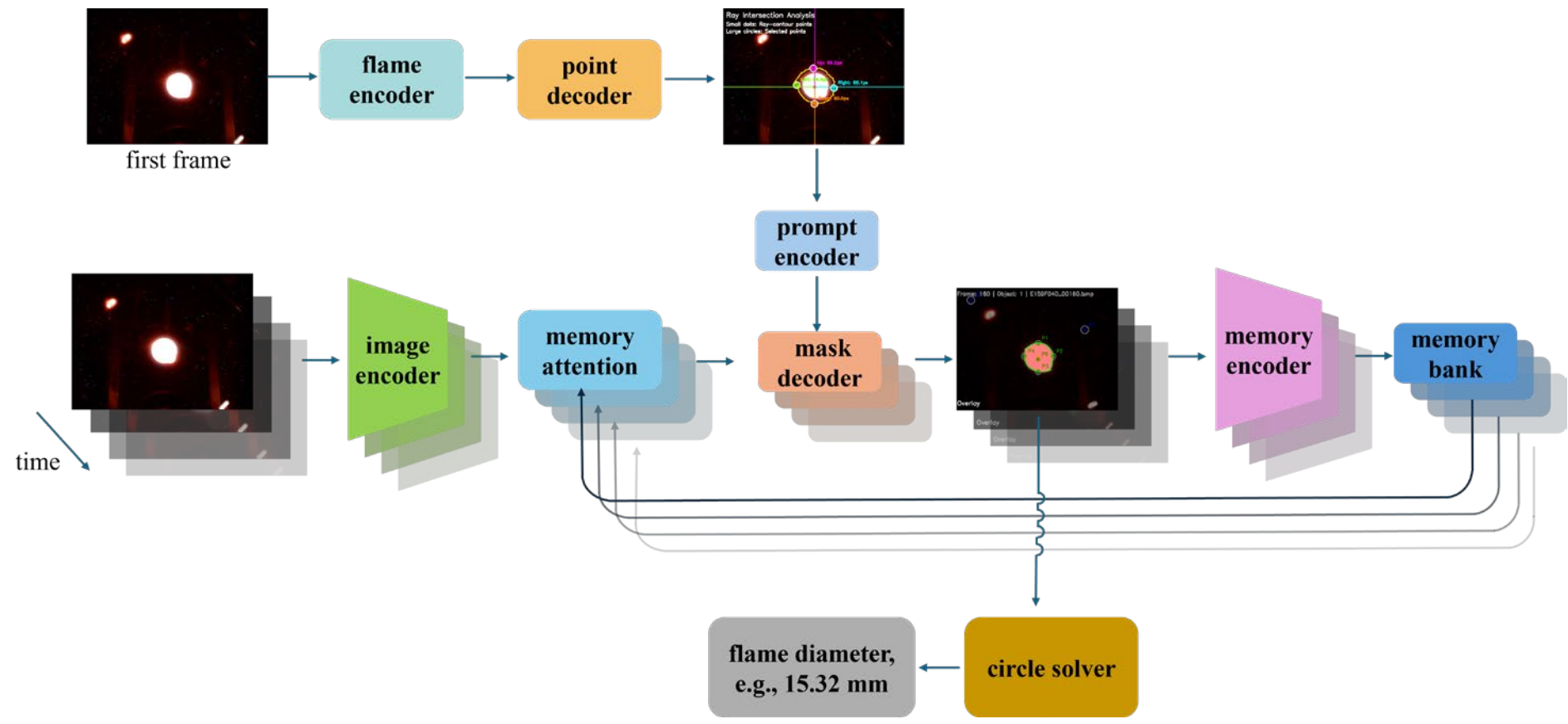

**Figure 2.** The pipeline of the overall framework for automatic flame diameter measurements via flame segmentation achieved by incorporating the automated p-points into SAM2.

The subsequent section will discuss the details of how to extend SAM2 to complete flame segmentation and measure flame diameter for various flame images under different challenging conditions, as previously shown in Figure 1.

### *3.1 Circle solver*

Figure 3 shows an example of a case (*n*-heptane with $D_0$ = 1.93 mm) where a soot tail is present (also see Figure 1b), which affects the measurement of $D_f$. Figures 3a, 3b, and 3c depict the original flame image input, the generated mask of flame without any prompt, and the overlay of the generated flame mask with the original flame, respectively. The generated flame mask in Figure 3b indicates that the soot tail will be inevitably detected and segmented together with the flame body. Unfortunately, the ending of the soot tail has already exceeded the flame boundary itself; this excess should not be considered to measure the $D_f$. Note that almost half of the burning droplets test runs have a soot tail, as shown in Table 1. Therefore, an accurate circle fitting method should be determined to avoid incorrectly including the soot tail as part of the flame before obtaining a precise flame boundary segmented mask. Figure 3d displays the result of the traditional convex circle fitting method on the generated flame mask. This method incorrectly fits the generated flame mask, significantly overestimating the actual flame diameter because of the influence of the soot tail. To resolve this issue, a RANSAC-based circle fitting procedure [30] is used in this paper to obtain an accurate measurement of $D_f$. This procedure discards the end of the soot tail pixels as outliers and then returns the flame diameter based on the inliers. Figure 3e shows the results. To validate the accuracy of the RANSAC-based circle fitting method used in this paper, we obtained quantitative data by comparing the accuracy of different classical circle fitting methods based on the test runs discussed in Section 1.1, including Convex [31], Full Least Squares [32], KASA method [33], and RANSAC [30]. The results show that the corresponding average accuracy $[(1-\frac{1}{N}\sum_{i=1}^{N}\frac{|D_{i,f,meas}-D_{i,f,manual}|}{D_{i,f,manual}})\times 100\%$, where $D_{i,f,meas}$ is the measured flame diameter by the different methods; $D_{f,\ manual}$ is the manually measured flame diameters outlined in [29, 34], serving as the manual-reference measurement (m-measurement) flame diameter] are 88.32%, 92.85%, 92.86%, and 98.19% respectively. These results clearly show that RANSAC is the most accurate method for circle fitting. The reason for this improvement is that RANSAC circle fitting successfully eliminates the soot tail region and thus returns a more accurate determination of the flame diameter.

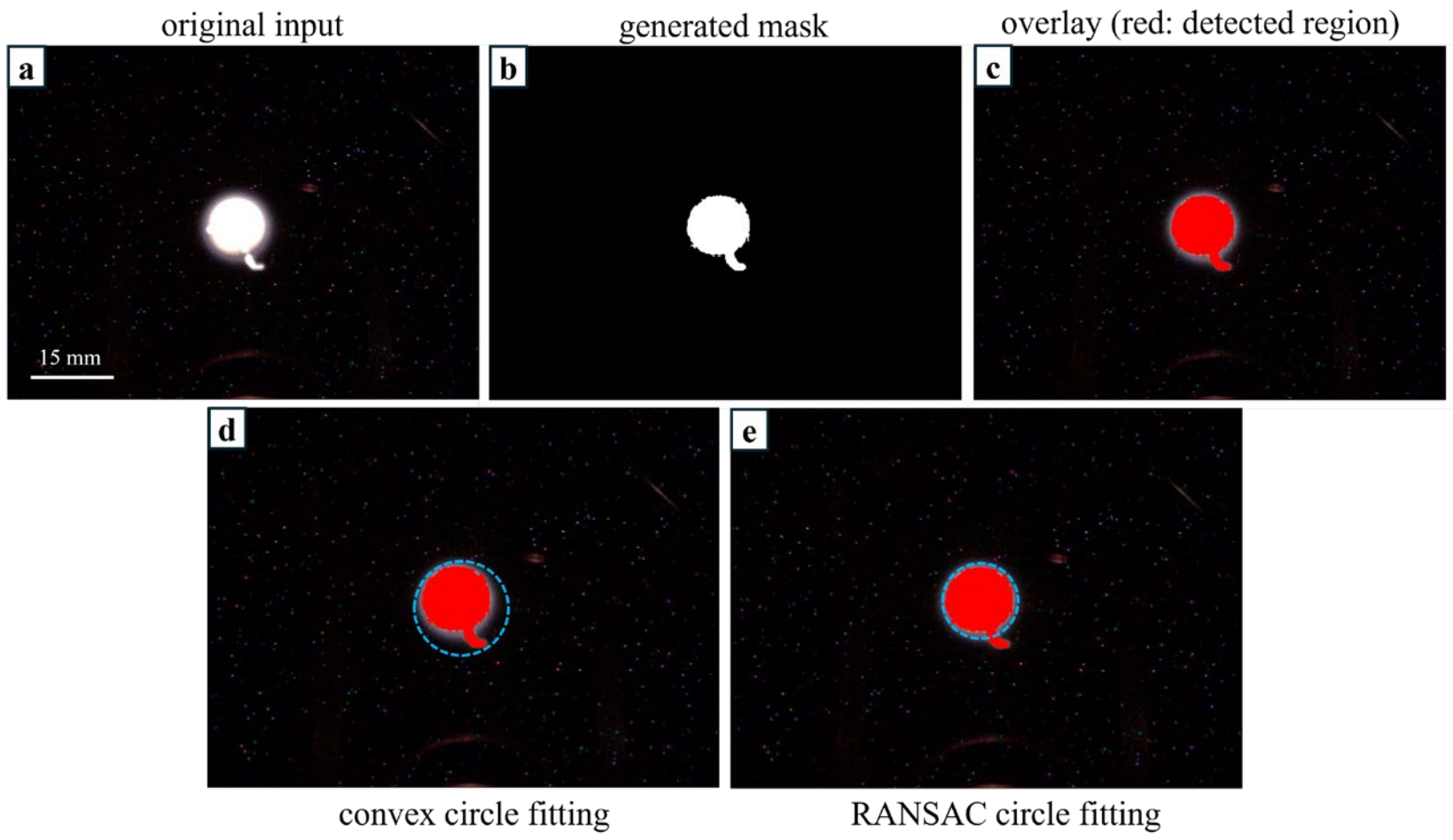

**Figure 3.** Demonstration of accurate circle fitting with RANSAC. **a.** an unprocessed flame image with a soot tail as the original input to the framework; **b.** generated flame mask (binary); **c.** mask overlay on original frame in which soot tail is also highlighted; **d.** convex circle fitting (blue dashed markers); **e.** final refined geometric least-squares fit (blue dashed markers) overlaid on the mask; note that the soot tail is now excluded from the fitted circle.

The details of how RANSAC works are briefly explained below. For a given two-dimensional mask in Figure 3b, a set of flame boundary points $\mathcal{D}=\{(x_i,y_i)\}_{i=1}^{N}$ can be extracted and used to estimate the circle parameters $\theta = (a, b, r)$ (center and radius). The flame boundary circle is then described as:

$$(x-a)^2+(y-b)^2=r^2, \qquad \theta=(a,b,r) \tag{1}$$

For each data point ($x_i$, $y_i$), define the geometric residual ε:

$$\varepsilon_i(\theta)=\left|\sqrt{(x_i-a)^2+(y_i-b)^2}-r\right| \tag{2}$$

A point is considered an inlier if $\varepsilon_i(\theta) \leq \tau$ (τ is a threshold distance in pixels). Then, RANSAC's objective is to search for the maximum number of inliers $\hat{\theta}$ [30] by:

$$\hat{\theta}=\arg\max_{\theta}\sum_{i=1}^{N}\mathbf{1}(\varepsilon_i(\theta)\leq\tau) \tag{3}$$

where $\mathbf{1}(\cdot)$ is the indicator function.

The number of iterations $k$ is defined by:

$$k = \frac{\log(1-p)}{\log(1-w^3)} \tag{4}$$

where $p$ is the confidence level, and $w$ is the estimated inlier ratio.

After RANSAC identifies the set of inliers, the optimized $\theta$ is computed, and the fitted flame diameter is obtained. This circle fitting method, which eliminates the effects of soot tails, is used throughout this paper to obtain precise flame diameter measurements.

Table 1
List of the number of tests analyzed in this work for various fuel systems

| Fuel Type | # of test runs analyzed | # of test runs with a soot tail | # of total images analyzed |
|---|---|---|---|
| $n$-Heptane | 16 | 11 | 5,956 |
| $n$-Decane | 14 | 5 | 6,256 |
| $n$-Octane | 15 | 7 | 6,325 |
| **Total** | **45** | **23** | **19,537** |

### *3.2 Effects of prompt-point selection on SAM2-based flame diameter measurement*

SAM2 has a core architecture of a prompt encoder (e.g., points, bounding boxes, or masks). To determine if the prompt encoder is necessary in the scenario involving an obscure flame boundary due to the "corona effect," some analyses are conducted to compare the results of SAM2 with and without some manual prompts. Figure 4a shows the representative flame images illustrating SAM2 segmentation of an $n$-heptane droplet ($D_0$ = 3.55 mm) with and without a manual p-point. $D_{f,\ meas}$ indicates the measured flame diameter by SAM2. $D_{f,\ manual}$ is manually measured flame diameter, serving as the m-measurements flame diameters. Figure 4b plots the results, comparing the flame diameter evolution as a function of normalized time $t/D_0^2$ between the SAM2 with manual prompts (red squares) or SAM2 without any prompts (black circles) and the m-measurements flame diameter. There are two important observations in the results plotted in Figure 4b. First, SAM2 with p-points improves the segmentation precision compared to the unprompted model. Second, there are still some slight differences between the $D_f$ measurements obtained from SAM2 with prompt guidance and the m-measurements values. The latter motivates us to perform further analyses to improve the model, which will be discussed in Section 3.4.

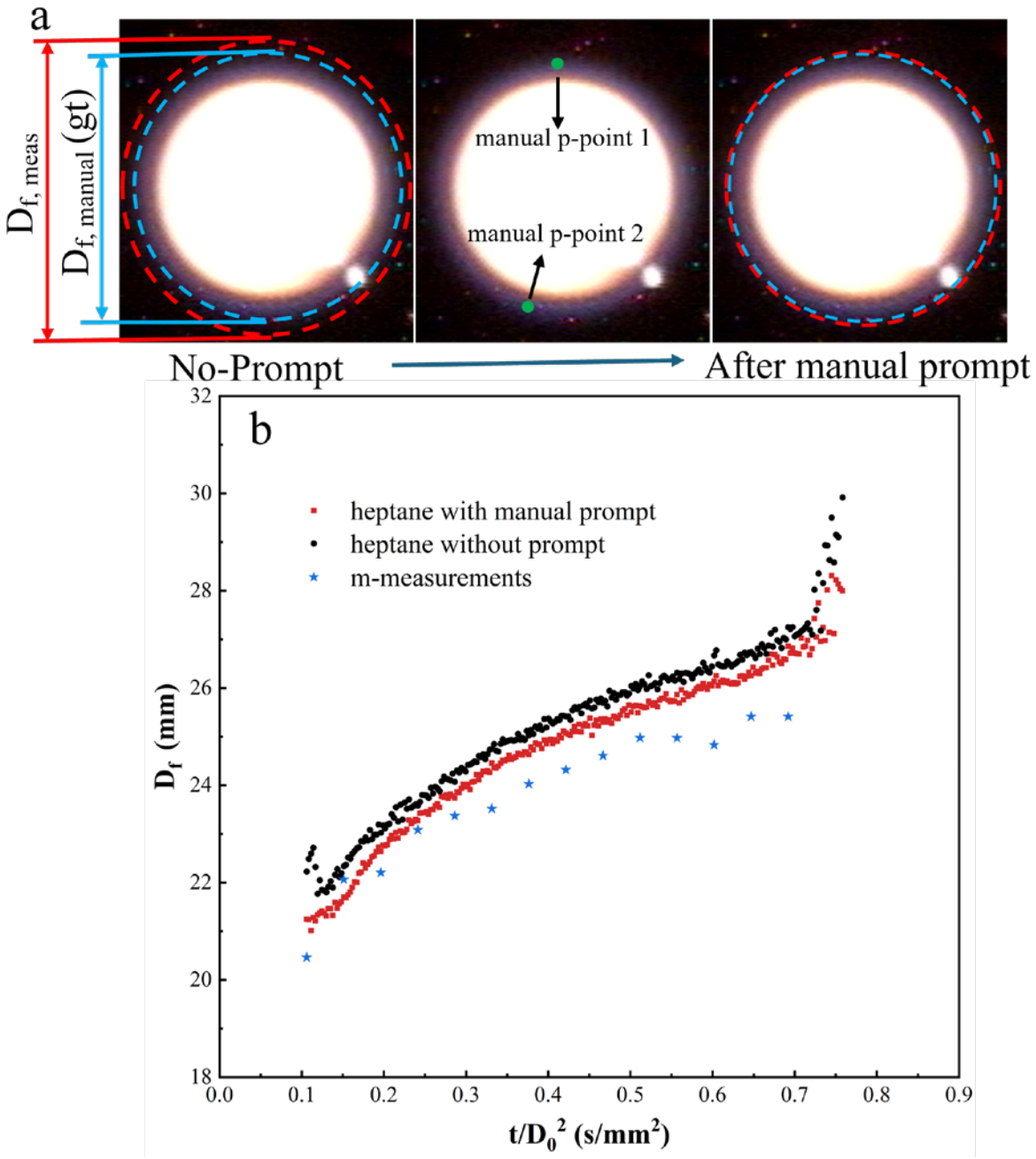

**Figure 4.** Effects of p-points on measuring the flame diameters using SAM2. **a.** Representative flame images illustrating SAM2 segmentation of an *n*-heptane droplet ($D_0$ = 3.55 mm) with and without a manual p-point. $D_{f,\ meas}$ indicates the measured flame diameter by SAM2. **b.** The comparison of flame diameter evolution as a function of normalized time $t/D_0^2$ between the SAM2 with manual prompts (red squares) or SAM2 without any prompts (black circles) and the m-measurements flame diameter.

There is another issue with the flame diameter measurements by SAM2 without prompts: Figure 5 highlights a failure case of SAM2 without any prompts being supplied. In this case, some reflections from the combustion chamber walls are erroneously segmented as flame (red overlay), resulting in fake flame diameter estimates exceeding 100 mm. In contrast, adding a few manual p-points effectively suppresses this artifact, aligning the SAM2 measurement with the m-measurements data.

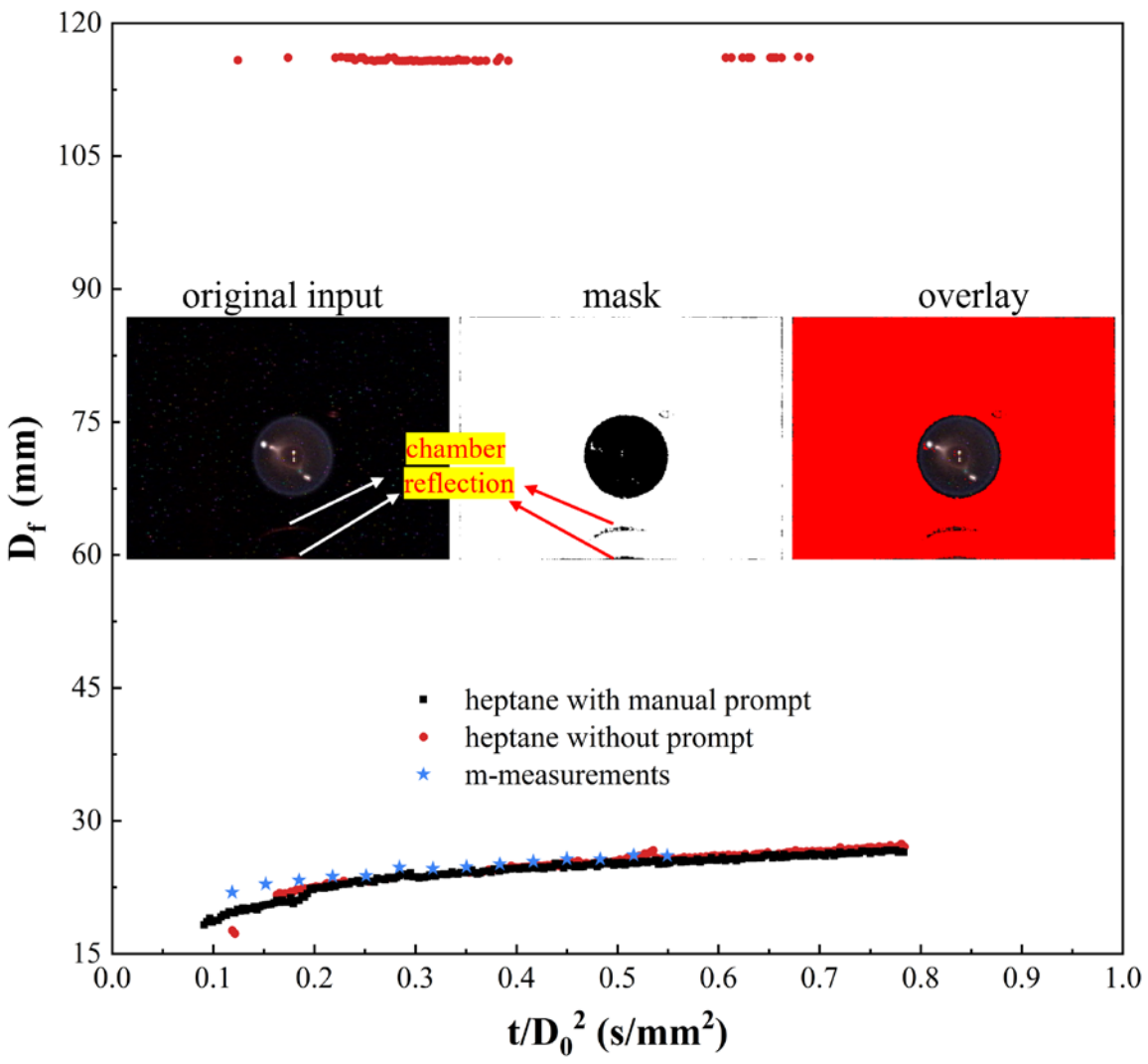

**Figure 5.** Representative failure case of SAM2 segmentation without manual prompts of a *n*-heptane droplet ($D_0$ = 3.47 mm). The chamber reflection is misdetected as the flame region and thus overestimates the flame diameter, exceeding 100 mm (the arrows indicate the chamber reflection, and the artifact SAM2 segmentation is shown in the red region, overlaid on the original input and mask).

### *3.3 Prompt-point placement strategy for flame segmentation*

Based on the above discussion, it is obvious that the p-points are necessary for accurately determining the flame boundary and suppressing the chamber reflection artifact. However, it is unclear whether the deployment and number of the p-points affect the accuracy of the flame diameter measurements via SAM2. This section will address this point. Figure 6a shows the SAM2 segmentation results of a particular flame image for a burning *n*-heptane droplet ($D_0$ = 3.47 mm) with different numbers of manually prompted points (indicated by green points). The blue prompt represents the negative value required to eliminate the background. The red area with a green circle is the fitted circle based on the SAM2 segmentation. The logic for deploying the manual prompts is to use one centroid of the burning droplet, plus the increasing number of annulus p-points in Figure 6a, with the annulus p-points remaining symmetrically distributed. It clearly indicates that the annulus flame boundary of the burning droplet is detected more accurately as the number of manual p-points increases. The annulus flame boundary of the burning droplet has been segmented almost completely when the manual p-points reach 5. Figure 6b plots the corresponding temporal evolution of flame diameter $D_f$ of Figure 6a with different numbers of manual p-points (manual p-points) against the m-measurements. Increasing the number of p-points progressively improves the accuracy of SAM2 measurements, demonstrating that prompt density reduces segmentation bias. The measured flame diameter results for 5 and 7 manual p-points overlap almost exactly, and the results are highly accurate, with average accuracy percentages of 98.05% and 98.17%, respectively. These results indicate that five manual p-points (i.e., one centroid point plus four symmetric annulus p-points) are sufficient to obtain accurate measurements of flame diameter. Therefore, the deployment of the five manual prompts is used for the rest of the paper to achieve accurate flame segmentation.

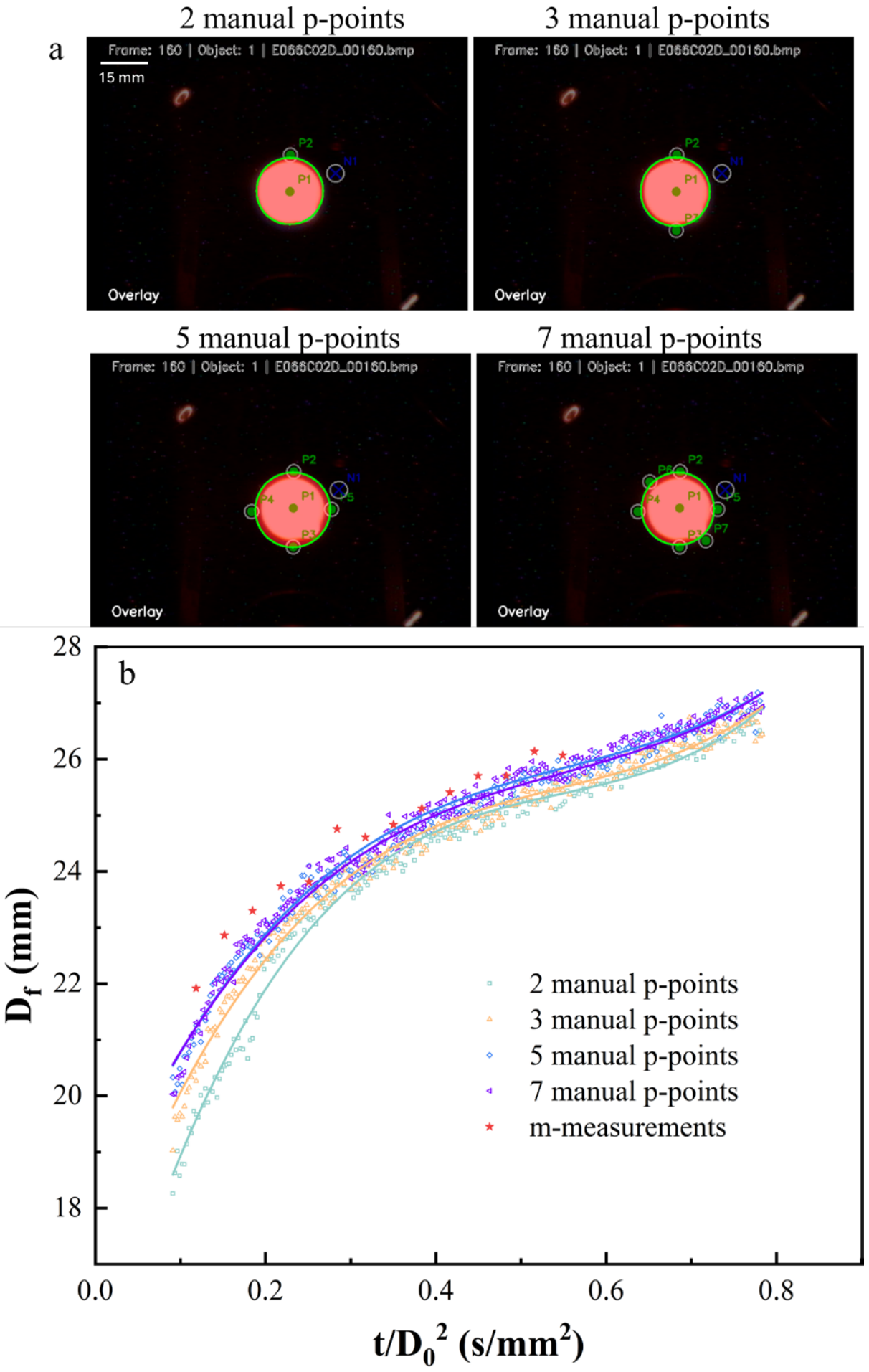

**Figure 6.** Effects of the deployment and number of p-points on $D_f$ measurement accuracy. **a.** Representative images showing SAM2 segmentation of a burning *n*-heptane droplet ($D_0$ = 3.47 mm) with different numbers of manual p-points (green points). **b.** The temporal evolution of flame diameter $D_f$ with different numbers of manual p-points (manual p-points) versus the normalized time $t/D_0^2$ against the m-measurements. Symbols represent actual flame diameter measurements; the solid lines are from the curve fitting of individual $D_f$ data points.

### *3.4 Automatic prompt-point generation framework*

Although appropriate deployment of manual p-points can yield precise flame segmentation, the manual prompt inevitably introduces subjective deviations and falls short of expectations for full automation. To achieve full automation of flame-diameter measurements and eliminate subjective bias in manual prompt selection, symmetric p-points should be generated automatically for SAM2-based flame segmentation. Figure 7 shows the automated procedure for generating symmetric p-points, which replaces manual selection in SAM2 flame segmentation. The original RGB input is converted to HSV color space, which decouples intensity (Value) from chromaticity (Saturation), thereby improving robustness against specular reflections and soot-induced color variations. Three binary masks are then constructed: (i) a brightness mask from the Value channel ($V > 25$); (ii) a saturation mask from the Saturation channel ($S > 15$); and (iii) a color-sum mask from the RGB channels ($R+G+B > 40$). These three masks are combined and refined using morphological closing (kernel size of 5×5) followed by opening (kernel size of 3×3) to suppress isolated noise.

The largest external contour is then extracted (yellow ‘circle’ in Figure 7), and the flame centroid is estimated using the minimum enclosing circle. Four orthogonal rays (up, down, left, right) are plotted starting from the centroid, and their intersections with the contour are identified. To reduce the influence of spurious protrusions, the intersection points are retracted toward the centroid by a constant shrink factor (0.9). These four symmetric points plus the centroid served as positive p-points for SAM2, while two additional negative p-points (a fixed background point and the darkest peripheral corner) are included to improve background discrimination.

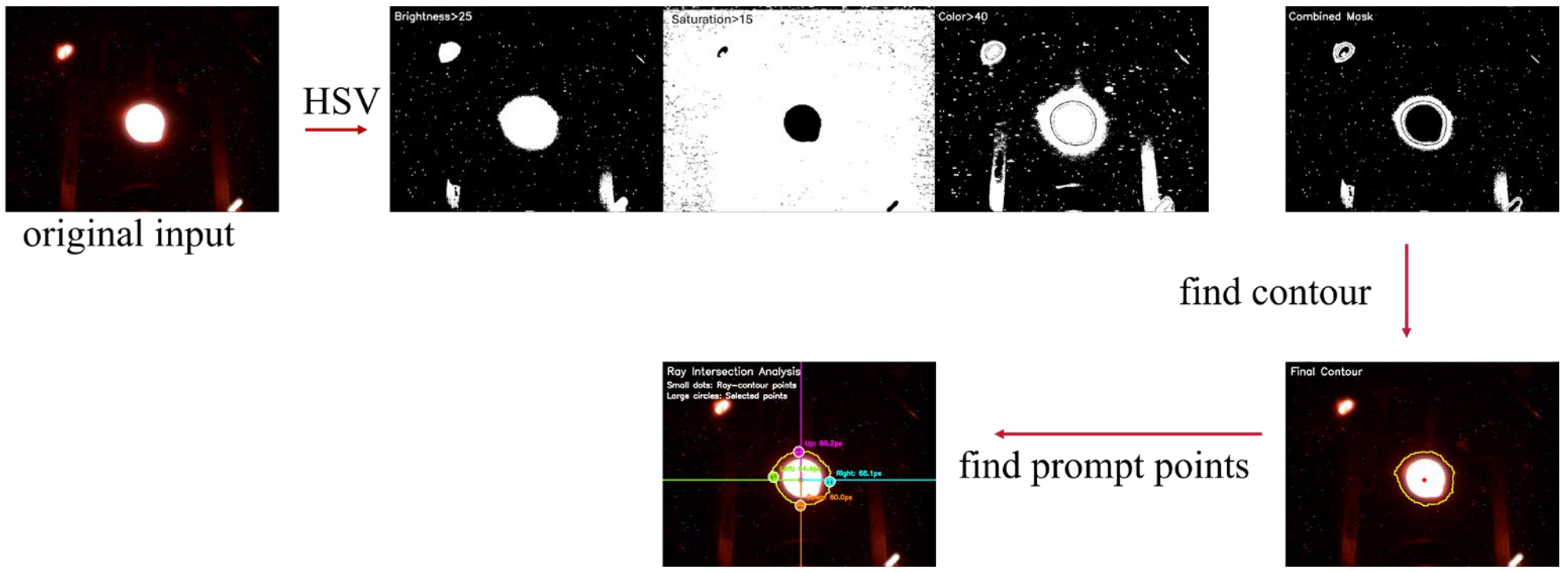

**Figure 7.** The pipeline for the automated generation of symmetric p-points for SAM2-based flame segmentation.

To further assess segmentation performance, we employ two standard overlap metrics: the Intersection-over-Union ($IoU = |A \cap B|/|A \cup B|$) [35] and the Dice coefficient ($Dice = 2|A \cap B|/(|A|+|B|)$) [36], where A is the segmentation area from the generated mask and B is the manual-reference measurements value. As shown in Figure 8, increasing the number of manual p-points gradually improves segmentation accuracy, which is consistent with previous discussions (also see Figure 6). Meanwhile, the automatically generated five-point strategy consistently outperforms the manual two-, three-, five-, and seven-point configurations. Specifically, the auto-prompt method maintains IoU values above 0.93 and Dice scores above 0.95 for most images analyzed for Figure 8, indicating both high accuracy and stability. These results confirm that the proposed auto-prompt strategy not only eliminates subjective variability but also provides a reliable alternative to labor-intensive manual point selection.

After establishing five p-points as the baseline strategy, the flame diameters are measured using both manually generated and automatically generated prompts. As shown in Figure 9, the two curves are highly consistent with each other and both align closely with the m-measurements. Importantly, the automatic approach slightly outperforms the manual strategy in terms of accuracy, with average agreement levels of 98.95% versus 98.05% for manual prompts. These results highlight that the auto-prompt method not only eliminates subjective intervention but also provides enhanced accuracy in flame diameter extraction.

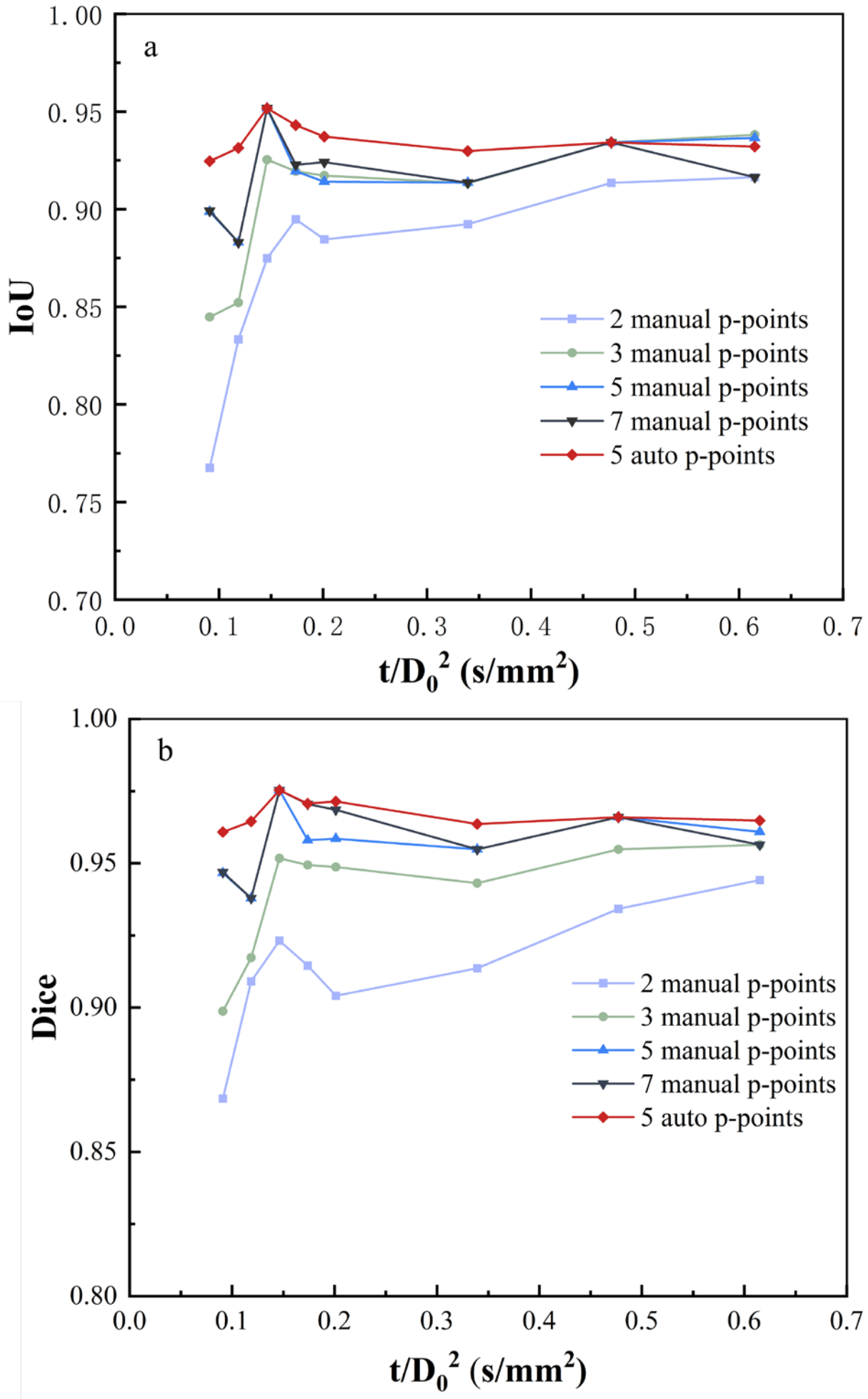

**Figure 8.** Quantitative evaluation of segmentation accuracy under different prompt strategies as a function of a normalized time $t/D_0^2$: **a.** Intersection-over-Union; **b.** Dice coefficient. Results show that using five automatically generated p-points (red line) consistently achieves accuracy comparable to, or better than, multiple manually selected p-points, thereby reducing manual prompt selection bias and improving reproducibility.

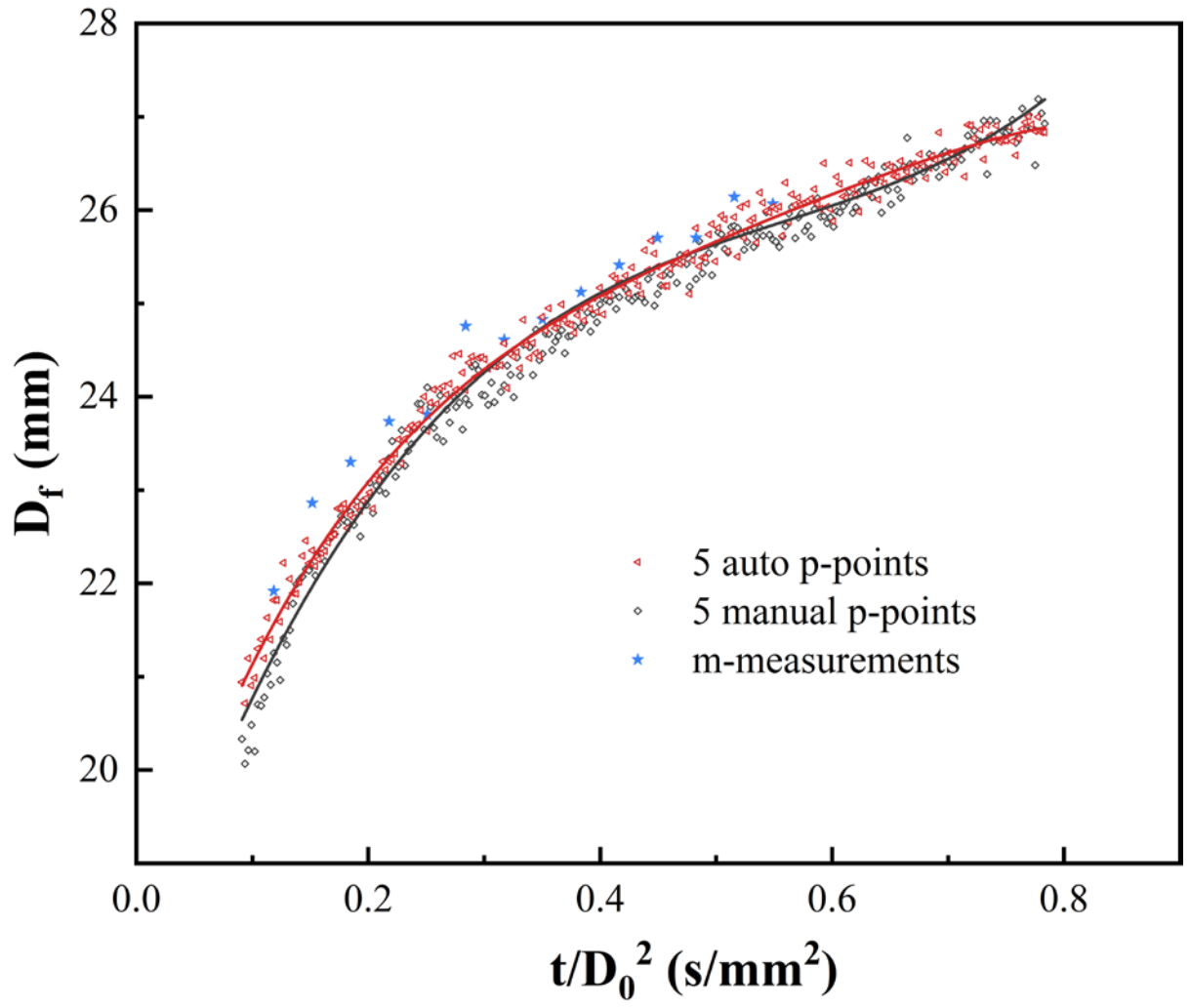

**Figure 9.** Comparison of $D_f$ measurements obtained from different strategies. Results are shown for five manual p-points (black), five automatically generated p-points (red), and m-measurements (blue). The close agreement between auto- and manual-prompt methods demonstrates that the proposed automatic strategy achieves equivalent accuracy while eliminating the need for manual intervention.

### *3.5 Video-based droplet tracking for drift compensation*

In FLEX-2 droplet combustion experiments, the slow drift of the droplet and flame, as shown in Figure 1c, strongly influences frame-by-frame image segmentation. Without tracking the flame centroid, the image model assumes a fixed flame position; consequently, drift causes the flame contour to shift relative to the generated mask. Even small pixel-level misalignments lead to unstable centroid detection and amplified fluctuations in the extracted flame diameter. To resolve this issue, the video model of SAM2 incorporates temporal tracking, dynamically updating the droplet position across frames. Figure 10a shows that the SAM2 video model accurately detects the centroid trajectories of the drifting flame, thereby eliminating the drift-induced fluctuations. On the other hand, the droplet centroid trajectories detected by the SAM2 image model concentrate mainly in one area. This does not correspond to the actual trajectory of the droplets during the drifting process, meaning that the deviation in droplet drift remains uneliminated. Such deviations originate from the positional variation of the flame centroid ($x_c$, $y_c$) and the uncertainty of the fitted flame radius $r$. The instantaneous deviation of the measured flame diameter from the error propagation principle can be expressed as:

$$\delta D_f = \frac{\partial D_f}{\partial x_c}\delta x_c + \frac{\partial D_f}{\partial y_c}\delta y_c + \frac{\partial D_f}{\partial r}\delta r \quad (5)$$

Since the flame diameter is typically defined as $D_f(x_c, y_c) = 2r(x_c, y_c)$, this relation simplifies to

$$\delta D_f \approx 2\,\delta r + \varepsilon_d \quad (6)$$

where $\varepsilon_d$ represents the drift-induced contribution caused by centroid displacement, which can be estimated as

$$\varepsilon_d = \sqrt{(\delta x_c)^2 + (\delta y_c)^2} \quad (7)$$

Consequently, when the SAM2 image model (SAM2 without temporal memory) is used, the lack of inter-frame consistency amplifies $\varepsilon_d$, leading to unstable diameter estimation.

In contrast, SAM2's video mode introduces a memory attention mechanism that establishes temporal correspondence between consecutive frames. This memory module constrains the spatial evolution of object masks, effectively suppressing $\delta x_c$ and $\delta y_c$. As a result, the measured $D_f$ becomes temporally smoother, ensuring more stable flame-diameter tracking during flame drifting.

The flame diameters measured by these two different SAM2 models also verify this truth. Figure 10b displays a comparison of the flame diameter as a function of normalized time ($t/D_0^2$) between the SAM2 video model and image model against the m-measurements. The correlation coefficients of the SAM2 image model and the video model are 0.776 and 0.97, respectively. In summary, this SAM2 video model stabilizes the centroid, suppresses drift-induced noise, and yields a much smoother evolution of $D_f$ that agrees well with m-measurements.

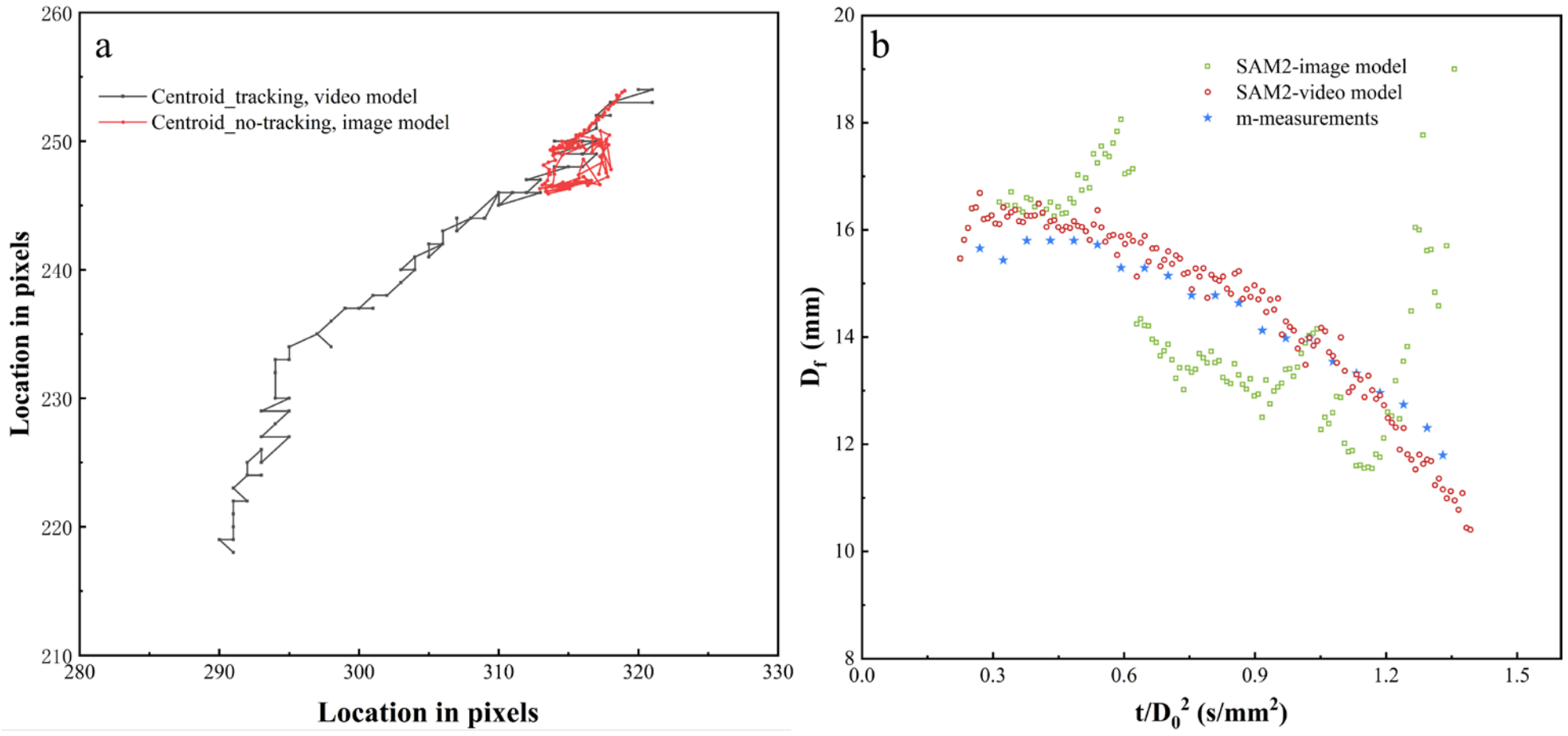

**Figure 10.** Using SAM2 video model to improve the $D_f$ measurement results due to droplet and flame drifting. **a.** Comparison of droplet centroid trajectories between the SAM2 video model and image model in pixel coordinates computed from segmentation masks. **b.** Comparison of flame diameter between the SAM2 video model and image model against the m-measurements.

### *3.6 Performance comparison with the conventional Hough circle detection method*

The performance of the newly developed auto-prompt measurement method in this work is further evaluated by comparing its measurements with both the conventional Hough circle detection method [13-15] and m-measurements. The Hough transform was adopted as the validation baseline because it is a classical and representative method for circular boundary detection. Figure 11 displays the flame diameter $D_f$ of a *n*-heptane droplet with $D_0$ = 1.93 mm, measured by the method developed in this work and the Hough circle detection method, plotted as a function of the normalized time $t/D_0^2$, with m-measurements included for reference. The detection output images are also displayed in Figure 11. The irregular white line in the inserted image in Figure 11 associated with the Hough circle detection is the flame boundary segmented by that method. As shown in the image, the flame boundary segmentation of the Hough method is non-circular, and the Hough circle detection obviously overestimates the flame boundary due to the blurry flame boundary caused by the "corona effects". In contrast, the red flame mask and the green fitting circle generated by the newly developed auto-prompt with SAM2 in this work show a precise flame boundary recognition. As a result, the flame diameters measured by the auto-prompt method reported here show good agreement with the m-measurements, and its average accuracy of 96.9% is much higher than that of the Hough circle detection (55.9%) because the Hough circle method misidentifies the flame boundary. Note that the RANSAC circle fitting method is used in both of the above circle segmentation methods to obtain the most accurate fitting diameter data.

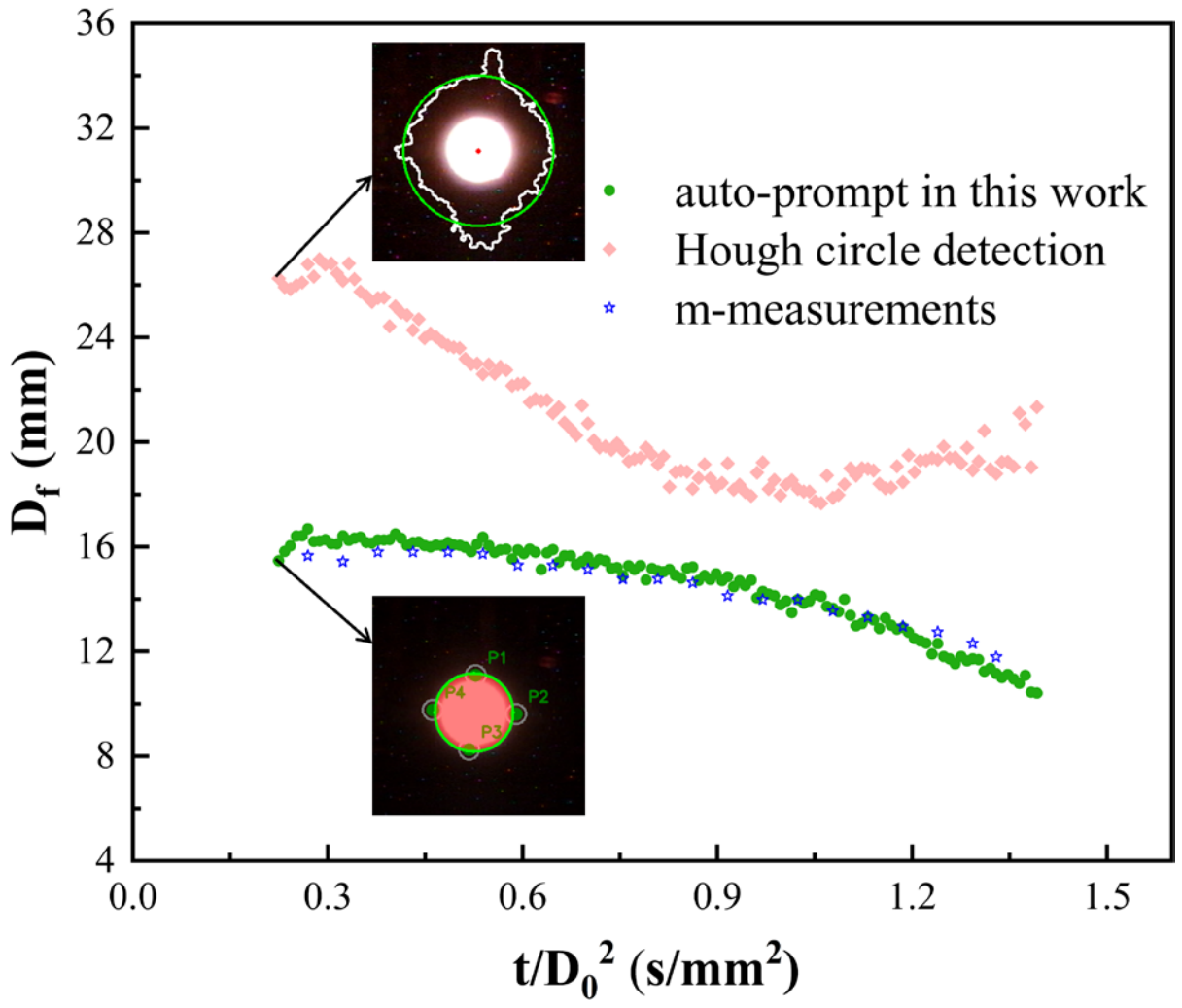

**Figure 11.** Comparison of flame diameter measurements obtained using the proposed auto-prompt method and a conventional Hough circle detection algorithm. The flame diameters $D_f$ of a *n*-heptane droplet with $D_0$ = 1.93 mm are plotted as a function of the normalized time $t/D_0^2$, with m-measurements included for reference.

### *3.7 Measurement uncertainty analysis*

The combined standard uncertainty of the automatically measured flame diameter is estimated by considering calibration $u_{cal}$, circle fitting $u_{fit}$, and the manual-reference measurements $u_{ref}$ contributions:

$$u_c^2(D_f) = u_{cal}^2 + u_{fit}^2 + u_{ref}^2 \tag{8}$$

where $u_{cal}^2 = (\sigma_L/L)^2 + (\sigma_P/P)^2$, $L$ = 93.2 mm and $P$ = 640 Pixels is the calibration factor. $u_L$ = 0.5 mm and $\sigma_P$ = 1 pixel are empirically assumed as the standard deviation; $u_{fit} = \sigma_f/D_f$ = 0.00625, $\sigma_f$ = 0.687 is the standard deviation by repeating the circle fitting 500 times for the same flame segmentation boundary, $D_f$ = 110 pixels is the average flame diameter by repeating the circle fitting 500 times for the flame boundary. $u_{ref} = \sigma_p/P = 0.085$ [29] is the uncertainty of the manual-reference measurements. As a result, the combined standard uncertainty of the $D_f$ measurements reported in this work is 8.54%.

## 4. Validation of automated flame diameter measurement for different fuels

Generally, the brightness of different types of fuels varies significantly in combustion because of the difference in each fuel's combustion chemistry. To verify the applicability of the framework developed in this paper, the diameters of flames for different fuels with varying droplet sizes are measured and reported here. Figure 12 summarizes the quantitative validation of the auto-prompt segmentation across different fuels. Figures 12a–c show the temporal evolution of flame diameter $D_f$ as a function of normalized time $t/D_0^2$ for *n*-heptane, *n*-decane, and *n*-octane droplets of various initial diameters, where the auto-prompt predictions (solid symbols) closely follow the m-measurements (open symbols, m-measurements). The overall consistency across all three fuels confirms that the newly developed framework can be applied to multiple fuels to accurately and automatically measure flame diameters, using a large dataset consisting of 19,537 images. As shown in Table 2, the newly developed framework only takes approximately 90 minutes to process all these images. However, if handled manually, it would take at least 20,000 minutes. This represents a significant improvement in efficiency of approximately 229×.

Table 2
Comparison of measurement time between m-measurements and the proposed workflow for various fuel systems

| Fuel Type | # of total images analyzed | m-measurements time (min) | workflow time in this study (min) |
|---|---|---|---|
| *n*-Heptane | 5,956 | 6,686 | 29 |
| *n*-Decane | 6,256 | 6,952 | 30 |
| *n*-Octane | 6,325 | 7,025 | 31 |
| **Total** | **19,537** | **20,663** | **90** |

Figure 12d plots the accuracy as a function of the initial droplet diameter $D_0$. A clear increasing trend is observed: smaller droplets ($D_0 \approx 1.5$–$2.0$ mm) yield accuracies around 96–97%, whereas larger droplets ($D_0 > 3.0$ mm) achieve nearly 99%. This increased accuracy as a result of increasing droplet size arises from three possible factors:

1. As discussed before, the blue flame at the end of the combustion stage (see Figure 1f) is relatively straightforward to measure. This blue flame period has the longest duration during droplet combustion and typically occurs for large droplets [3, 4, 22].
2. Larger droplets produce stronger luminous flames and clearer contours, reducing segmentation ambiguity.
3. Relative errors in diameter extraction decrease when the absolute droplet size is larger, since a given pixel-level uncertainty contributes less to the normalized diameter.

These results demonstrate that the auto-prompt strategy can be applied across different fuels and achieves higher accuracy for larger droplets, reaching an average accuracy of

98.1%.

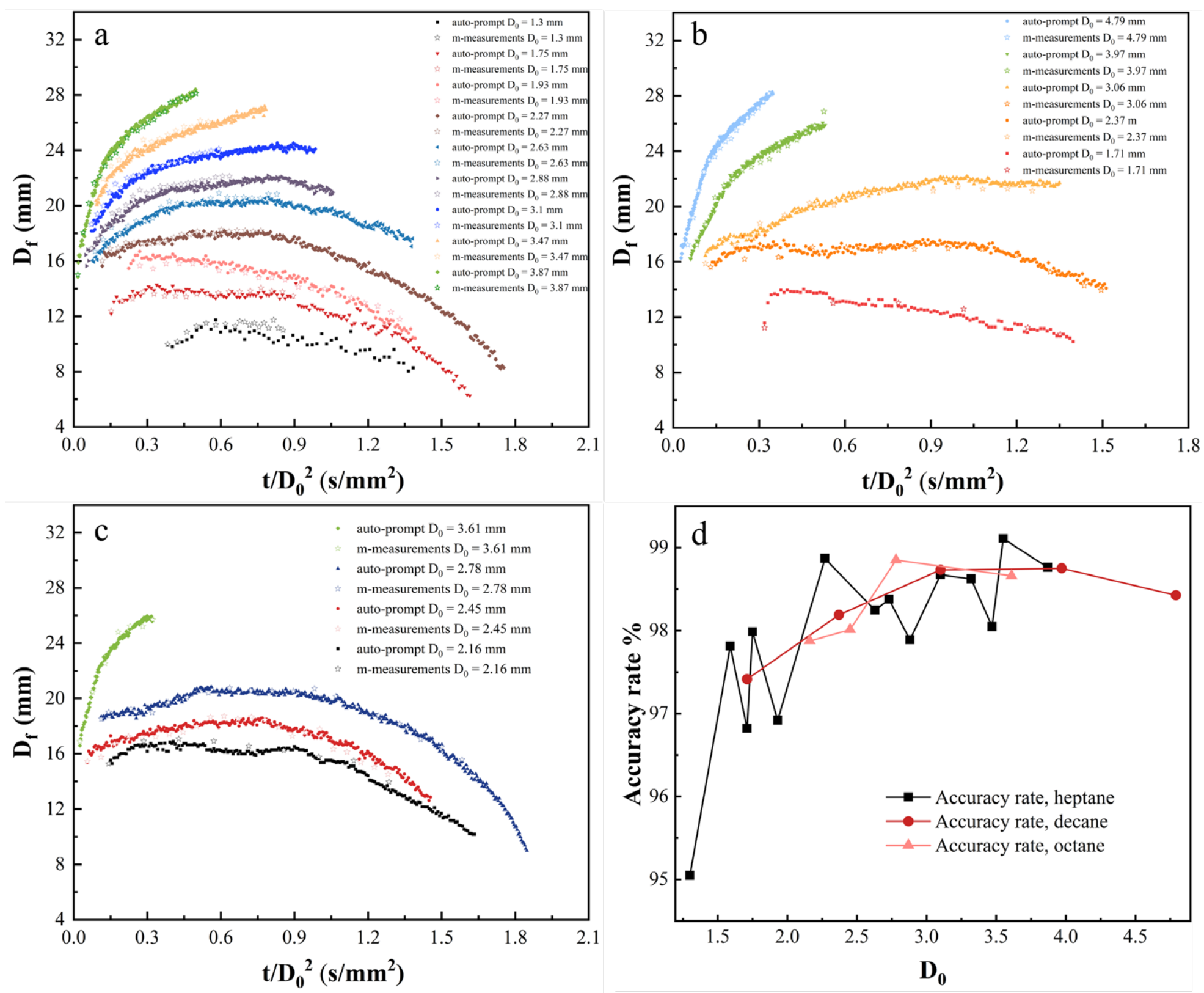

**Figure 12.** Comparison between auto-prompt results and m-measurements of flame diameters $D_f$ for **a.** *n*-heptane, **b.** *n*-decane, and **c.** *n*-octane droplets with different initial diameters $D_0$. The results show accurate $D_f$ measurements from SAM2 for all three fuels investigated, with increasing droplet size leading to improved accuracy. **d.** Average measurement accuracy as a function of $D_0$ for *n*-heptane, *n*-decane, and *n*-octane, indicating that the recognition accuracy rises with increasing droplet diameter.

## 5. Conclusion

In this work, we develop a SAM2-based digital workflow for automated flame segmentation and diameter measurement in droplet combustion experiments under microgravity environments. In this framework, a RANSAC-based circle-fitting method accurately suppresses the soot-tail-induced measurement bias by identifying and excluding outlier pixels, thereby significantly improving the accuracy of flame-diameter measurements under soot-tail conditions. An automatic prompt-generation framework is also incorporated into SAM2 to fully automate droplet flame diameter measurements and eliminate subjectivity bias associated with manual prompt generation. The accuracy of the automatic prompt generation (98.95%) is higher than that of the manual prompt generation (98.05%). Additionally, the SAM2 video model incorporates temporal tracking and dynamically updates the flame position frame-by-frame, thereby eliminating fluctuations caused by flame drift and significantly enhancing measurement stability. The average accuracy of flame diameter measurements using this new framework is 96.9%, showing a significant improvement over the traditional circular Hough transform (55.9%). In addition to accuracy evaluation, an uncertainty analysis is conducted to characterize the metrological reliability of the proposed digital measurement workflow. The combined standard uncertainty is estimated to be 8.54%.

Finally, this automatic prompt-point SAM2-based framework has been shown to have wide applicability across fuels, including *n*-heptane, *n*-decane, and *n*-octane, by measuring flame diameters from 19,537 images at different initial droplet sizes for each fuel. The results also show that the measurement accuracy increases with the initial droplet diameter. The proposed workflow also provides a major improvement in measurement efficiency. Processing the complete dataset required only a small fraction of the time needed for manual measurements, corresponding to an efficiency improvement of approximately 229×.

This fully automated flame-diameter measurement framework provides a generalizable, scalable solution for flame diagnostics and can be extended to other combustion-imaging applications that require robust segmentation and geometric quantification. It also makes the framework applicable to a wider range of image-based real-world measurement tasks, including feature extraction from low-contrast target boundary images, dynamic object tracking, and automated analysis of large visual datasets.

## CRediT authorship contribution statement

**Minghui Xu:** Formal analysis, Methodology, Writing – original draft. **Chaoyi Zhou:** Methodology. **Aaron P. Cecil:** Writing – original draft. **Xi Liu:** Writing – original draft. **Siyu Huang:** Writing – review & editing. **Yuhao Xu:** Conceptualization, Funding acquisition, Project administration, Supervision, Writing – review & editing.

## Funding

This work was supported in part by NSF Grant No. CBET-2349193.

## Declaration of competing interest

The authors declare that they have no known competing financial interests or personal relationships that could have appeared to influence the work reported in this paper.

## Data availability

The original unprocessed color images are publicly available at the Physical Sciences Informatics (PSI) website maintained by NASA [13]. Because the raw images are already publicly available through NASA PSI, they are not redistributed here. This present work is an analysis of these data. The processed data generated in this study are available from the corresponding authors upon reasonable request.